\documentclass[11pt,a4paper]{article}
\usepackage[hyperref]{emnlp2020}
\usepackage{times}
\usepackage{latexsym}

\usepackage{microtype}

\usepackage{graphicx}
\usepackage{adjustbox}
\usepackage{algorithm, algorithmic}
\usepackage{color}
\usepackage{array}
\usepackage{amsmath}
\usepackage{subfigure}
\usepackage{multirow}
\usepackage{textcomp}

\aclfinalcopy 
\def\aclpaperid{292} 

\title{Active Testing: An Unbiased Evaluation Method for Distantly Supervised Relation Extraction}

\author{Pengshuai Li$^1$, Xinsong Zhang$^2$, Weijia Jia$^{3,1}$\thanks{~~Corresponding author: jiawj@bnu.edu.cn.} \and Wei Zhao$^{4}$ \\
$^1$Dept. of CSE, Shanghai Jiao Tong University, Shanghai, China \quad
$^2$ByteDance AI Lab\\
$^3$Institute of AI \& Future Networks, Beijing Normal University (Zhuhai) \& UIC, PR China \\
$^4$American University of Sharjah, Sharjah, United Arab Emirates \\
{\tt pengshuai.li@sjtu.edu.cn \quad  zhangxinsong.0320@bytedance.com} \\
{\tt jiawj@bnu.edu.cn \quad wzhao@aus.edu} \\
}

\date{}

\begin{document}
\maketitle
\begin{abstract}
Distant supervision has been a widely used method for neural relation extraction for its convenience of automatically labeling datasets. However, existing works on distantly supervised relation extraction suffer from the low quality of test set, which leads to considerable biased performance evaluation. These biases not only result in unfair evaluations but also mislead the optimization of neural relation extraction. To mitigate this problem, we propose a novel evaluation method named active testing through utilizing both the noisy test set and a few manual annotations. Experiments on a widely used benchmark show that our proposed approach can yield approximately unbiased evaluations for distantly supervised relation extractors.
\end{abstract}

\section{Introduction}
\label{sec:intro}

Relation extraction aims to identify relations between a pair of entities in a sentence. It has been thoroughly researched by supervised methods with hand-labeled data. To break the bottleneck of manual labeling, distant supervision~\cite{mintz2009distant} automatically labels raw text with knowledge bases. It assumes that if a pair of entities have a known relation in a knowledge base, all sentences with these two entities may express the same relation. Clearly, the automatically labeled datasets in distant supervision contain amounts of sentences with wrong relation labels. However, previous works only focus on wrongly labeled instances in training sets but neglect those in test sets. Most of them estimate their performance with the held-out evaluation on noisy test sets, which will yield inaccurate evaluations of existing models and seriously mislead the model optimization. As shown in Table~\ref{table:evaluations}, we compare the results of held-out evaluation and human evaluation for the same model on a widely used benchmark dataset NYT-10~\cite{riedel2010modeling}. The biases between human evaluation and existing held-out evaluation are over 10\%, which are mainly caused by wrongly labeled instances in the test set, especially false negative instances.

\begin{table}[htbp]
\centering
\resizebox{\linewidth}{!}{
\begin{tabular}{cccc}
\hline
Evaluations & P@100 & P@200 & P@300 \\
\hline
Held-out Evaluation  & 83 & 77 & 69\\
Human Evaluation  & 93({\footnotesize +10}) & 92.5({\footnotesize +15.5}) & 91({\footnotesize +22}) \\
\hline
\end{tabular}}
\caption{The Precision at top K predictions (\%) of the model~\citeauthor{lin2016neural}~\shortcite{lin2016neural} upon held-out evaluation and human evaluation on NYT-10. Results are obtained by our implementations.}
\label{table:evaluations}
\end{table}

A false negative instance is an entity pair labeled as non-relation, even if it has at least one relation in reality.  This problem is caused by the incompleteness of existing knowledge bases. For example, over 70\% of people included in Freebase have no place of birth~\cite{dong2014knowledge}. From a random sampling, we deduce that about 8.75\% entity pairs in the test set of NYT-10 are misclassified as non-relation.\footnote{We randomly selected 400 entity pairs from the test set, in which 35 are misclassified as non-relation.} Clearly, these mislabeled entity pairs yield biased evaluations and lead to inappropriate optimization for distantly supervised relation extraction. 

In this paper, we propose an active testing approach to estimate the performance of distantly supervised relation extraction. Active testing has been proved effective in evaluating vision models with large-scale noisy datasets~\cite{Nguyen2018ActiveTA}. In our approach, we design an iterative approach, with two stage per iteration: vetting stage and estimating stage. In the vetting stage, we adopt an active strategy to select batches of the most valuable entity pairs from the noisy test set for annotating. In the estimating stage, a metric estimator is proposed to obtain a more accurate evaluation. With a few vetting-estimating iterations, evaluation results can be dramatically close to that of human evaluation by using limited vetted data and all noisy data. Experimental results demonstrate that the proposed evaluation method yields approximately unbiased estimations for distantly supervised relation extraction.

\section{Related Work}
\label{sec:rela}
Distant supervision~\cite{mintz2009distant} was proposed to deal with large-scale relation extraction with automatic annotations. A series of studies have been conducted with human-designed features in distantly supervised relation extraction~\cite{riedel2010modeling, surdeanu2012multi,takamatsu2012reducing,angeli2014combining, han2016global}. In recent years, neural models were widely used to extract semantic meanings accurately without hand-designed features~\cite{zeng2015distant, lin2017neural, zhang2018multi}. Then, to alleviate the influence of wrongly labeled instances in distant supervision, those neural relation extractors integrated techniques such as attention mechanism~\cite{lin2016neural,han2018hierarchical,huang2019self}, generative adversarial nets~\cite{qin2018dsgan, li2019gan}, and reinforcement learning~\cite{feng2018reinforcement, qin2018robust}. However, none of the above methods pay attention to the biased and inaccurate test set. Though human evaluation can yield accurate evaluation results~\cite{zeng2015distant,alt2019fine}, labeling all the instances in the test set is too costly.

\section{Task Definition}
\label{sec:task}
In distant supervision paradigm, all sentences containing the same entity pair constitute a bag. Researchers train a relation extractor based on bags of sentences and then use it to predict relations of entity pairs. Suppose that a distantly supervised model returns confident score\footnote{Confident scores are estimated probabilities for relations.} $s_i=\{s_{i1}, s_{i2} \dots s_{ip}\}$ for entity pair $i \in \{1 \dots N\}$, where $p$ is the number of relations, $N$ is the number of entity pairs, and $s_{ij} \in (0,1)$. $y_i=\{y_{i1}, y_{i2} \dots y_{ip}\}$ and $z_i=\{z_{i1}, z_{i2} \dots z_{ip}\}$ respectively represent automatic labels and true labels for entity pair $i$, where $y_{ij}$ and $z_{ij}$ are both in $\{0, 1\}$\footnote{An entity pair may have more than one relations.}.

In widely used held-out evaluation, existing methods observe two key metrics which are precision at top K ($P@K$) and Precision-Recall curve ($\textit{PR} \ curve$). To compute both metrics, confident score for all entity pairs are sorted in descending order, which is defined as $s'=\{s'_1, s'_2 \dots s'_P\}$ where $P=Np$. Automatic labels and true labels are denoted as $y'=\{y'_1, \dots, y'_P\}$ and $z'=\{z'_1, \dots, z'_P\}$. In summary, $P@K$ and $R@K$ can be described by the following equations,
\begin{equation}
    P@K\{z'_1 \dots z'_P\}=\frac{1}{K}\sum_{i \le K}z'_i
\end{equation}
\begin{equation}
    R@K\{z'_1 \dots z'_P\}=\frac{\sum_{i\le K}z'_i}{\sum_{i\le P}z'_i}
\end{equation}
Held-out evaluation replaces $z'$ with $y'$ to calculate $P@K$ and $R@K$, which leads to incorrect results obviously.

\section{Methodology}
\label{sec:meth}
In this section, we present the general framework of our method. A small random sampled set is vetted in the initial state. In each iteration there are two steps: 1) select a batch of entity pairs with a customized vetting strategy, label them manually, and add them to the vetted set; 2) use a new metric estimator to evaluate existing models by the noisy set and the vetted set jointly. After a few vetting-evaluating iterations, unbiased performance of relation extraction is appropriately evaluated. In summary, our method consists of two key components: a vetting strategy and a metric estimator.

\subsection{Metric Estimator}
Our test set consists of two parts: 1) a noisy set $U$ in which we only know automatic label $y'_i$; 2) a vetted set $V$ in which we know both automatic label $y'_i$ and manual label $\tilde z'_i$. We treat the true label $z'_i$ as a latent variable and $\tilde z'_i$ is its observed value. The performance evaluation mainly depends on the estimation of $z'_i$. In our work, we estimate the probability as
\begin{equation}
    p(z'_i)=\prod_{i\in U}p(z'_i|\Theta)\prod_{i\in V}\delta(z'_i=\tilde z'_i)
\end{equation}
where $\Theta$ represents all available elements such as confident score, noisy labels and so on. We make the assumption that the distribution of true latent labels is conditioned on $\Theta$.

Given posterior estimates $p(z'_i|\Theta)$, we can compute the expected performance by replacing the true latent label by its probability. Then, the precision and recall equations can be rewritten as
\begin{equation}
    E[P@K]  =  \frac{1}{K}(\sum_{i\in V_K} \tilde z'_i  +  \sum_{i\in U_K}p(z'_i  =  1|\Theta))
\end{equation}
\begin{equation}
    E[R@K]=\frac{\sum_{i\in V_K} \tilde z'_i + \sum_{i\in U_K}p(z'_i=1|\Theta)}{\sum_{i\in V} \tilde z'_i + \sum_{i\in U}p(z'_i=1|\Theta)}
\end{equation}
where $U_K$ and $V_K$ denote the unvetted and vetted subsets of $K$ highest-scoring examples in the total set $U \cup V$.

To predict the true latent label $z'_i$ for a specific relation, we use noisy label $y'_i$ and confident score $s'_i$. This posterior probability can be derived as (see appendix for proof)
\begin{equation}
\begin{split}
 p(z'_i|y'_i, s'_i) = 
     \frac{p(y_{jk}|z_{jk})p(z_{jk}|s_{jk})}{\sum_v p(y_{jk}|z_{jk}=v)p(z_{jk}=v|s_{jk})}
\end{split}
\label{equ:pzi}
\end{equation}
where $v \in \{0,1\}$. $s_{jk}, y_{jk}, z_{jk}$ are the corresponding elements of $s'_{i}, y'_i, z'_i$ before sorting confident score. Given a few vetted data, we fit $p(y_{jk}|z_{jk})$ by standard maximum likelihood estimation (counting frequencies). $p(z_{jk}|s_{jk})$ is fitted by using logistic regression. For each relation, there is a specific logistic regression function to fit. 

\subsection{Vetting Strategy}
In this work, we apply a strategy based on $maximum \  expected \  model\ change$(MEMC)~\cite{settles2009active}. The vetting strategy is to select the sample which can yield a largest expected change of performance estimation. Let $E_{p(z'|V)}Q$ be the expected performance based on the distribution $p(z'|V)$ estimated from current vetted set $V$. After vetting example $i$ and updating that estimator, it will become $E_{p(z'|V,z'_i)}Q$. The change caused by vetting example $i$ can be written as
\begin{equation}
    \Delta_i(z'_i) = \vert E_{p(z'|V)}Q - E_{p(z'|V,z'_i)}Q \vert
\end{equation}
For precision at top K, this expected change can be written as 
\begin{equation}
    E_{p(z'_i|V)}[\Delta_i(z'_i)] = \frac{2}{K}p_i(1-p_i)
\label{equ:epz}
\end{equation}
where $p_i=P(z'_i=1|\Theta)$. For the PR curve, every point depends on $P@K$ for different $K$. Thus, this vetting strategy is also useful for the PR curve.

With this vetting strategy, the most valuable data is always selected first. Therefore, vetting budget is the only factor controlling the vetting procedure. In this approach, we take it as a hyper parameter. When the budget is used up, the vetting stops. The procedure is described in Algorithm~\ref{alg:active}.

\begin{algorithm}
\caption{Active Testing Algorithm}
\label{alg:active}
\begin{algorithmic}[1]
\REQUIRE unvetted set $U$, vetted set $V$, vetting budget $T$, vetting strategy \emph{VS}, confident score $S$, estimator $p(z')$
\WHILE{$T>0$}
    \STATE select a batch of items $B \in U$ with vetting strategy \emph{VS}
    \STATE vet B and get manual label $\tilde z'$
    \STATE U=U\textminus B, V=V$\cup$B
    \STATE fit $p(z')$ with $U,V,S$
    \STATE T=T\textminus $\vert B \vert$
\ENDWHILE
\end{algorithmic}
\end{algorithm}

\section{Experiment}
\label{sec:expe}
We conduct sufficient experiments to support our claims; 1) The proposed active testing is able to get more accurate results by introducing very few manual annotations. 2) The held-out evaluation will misdirect the optimization of relation extraction, which can be further proved through re-evaluation of eight up-to-date relation extractors.

\subsection{Experimental Setting}

\paragraph{Dataset.}
Our experiments are conducted on a widely used benchmark NYT-10~\cite{riedel2010modeling} and an accurate dataset named NYT-19, which contains 500 randomly selected entity pairs from the test set of NYT-10. It contains 106 positive entity pairs and 394 negative entity pairs, in which 35 entity pairs are false negative. NYT-19 has been well labeled by NLP researchers.

\paragraph{Initialization.}
We use PCNN+ATT~\cite{lin2016neural} as baseline relation extractors. To be more convincing, we provide the experimental results of BGRU+ATT in the appendix. The initial state of vetted set includes all the positive entity pairs of the test set in NYT-10 and 150 vetted negative entity pairs. The batch size for vetting is 20 and the vetting budget is set to 100 entity pairs.

\subsection{Effect of Active Testing}

\begin{figure}[ht]
\centering
\includegraphics[width=0.98\linewidth]{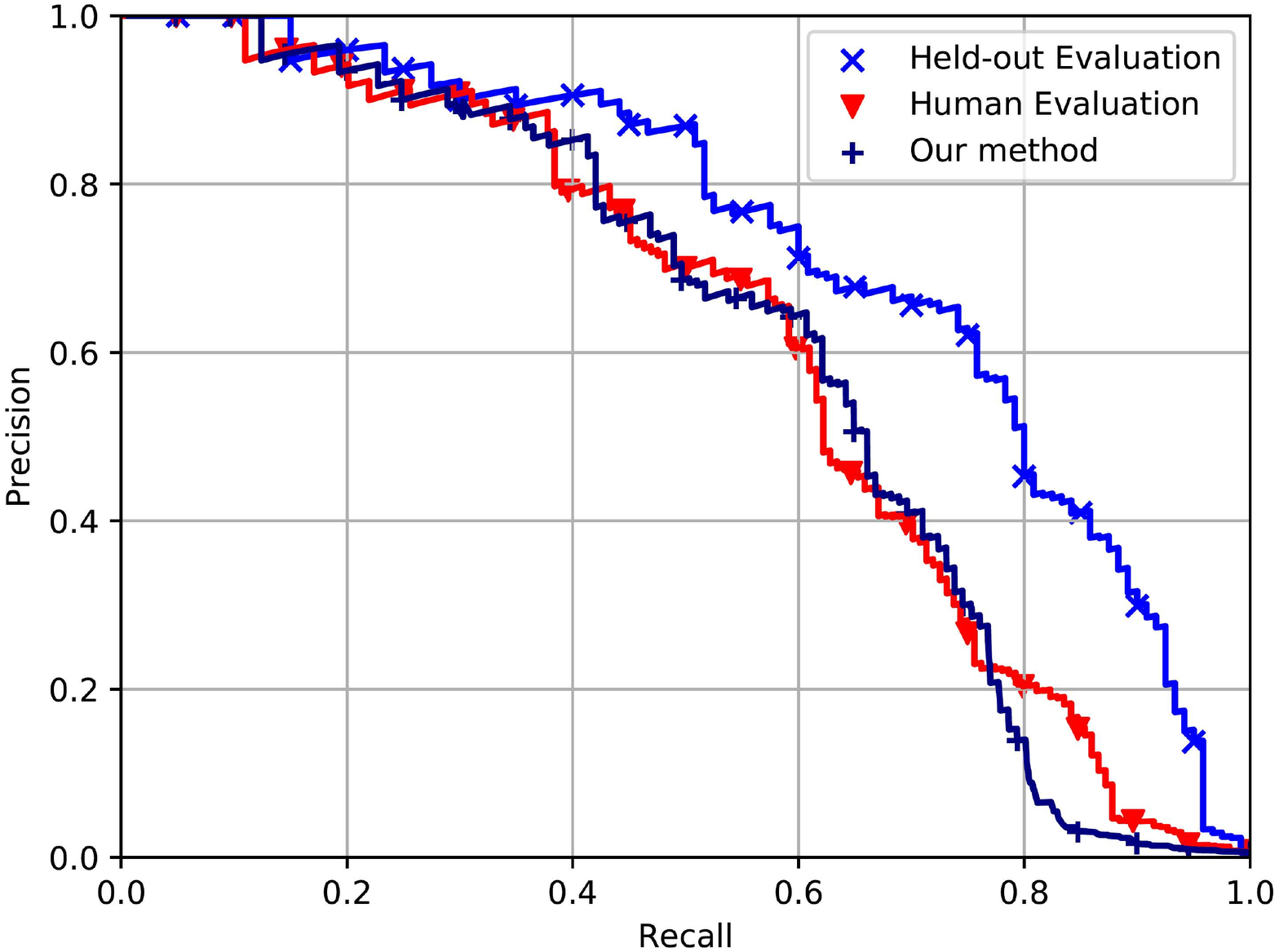}
\caption{The PR curve of PCNN+ATT on NYT-19.}
\label{fig:pcnn_PR}
\end{figure}

We evaluate the performance of PCNN+ATT with held-out evaluation, human evaluation and our method. The results are shown in Table~\ref{table:result_topk}, and Figure~\ref{fig:pcnn_PR}. Due to high costs of manual labeling for the whole test set, we use the PR-curve on NYT-19 to simulate that on NYT-10.

\begin{table}[htbp]
\centering
\resizebox{\linewidth}{!}{
\begin{tabular}{ccccc}
\hline
Model & Evaluations & P@100 & P@200 & P@300 \\
\hline
\multirow{3}{*}{PCNN+ATT}& Held-out Evaluation  & 83 & 77 & 69\\
& Our method & 91.2 & 88.4 & 83.4 \\
& Human Evaluation  & 93 & 92.5& 91 \\
\hline
\end{tabular}}
\caption{The Precision at top K predictions (\%) of PCNN+ATT upon held-out evaluation, our method and human evaluation on NYT-10.}
\label{table:result_topk}
\end{table}

To measure the distance between two curves, we sample 20 points equidistant on each curve and calculate the Euclidean distance of the two vectors. In this way, our method gets the distances 0.17 to the curve of human evaluation while corresponding distances for held-out evaluation is 0.72. We can observe that 1) The performance biases between manual evaluation and held-out evaluation are too significant to be neglected. 2) The huge biases caused by wrongly labeled instances are dramatically alleviated by our method. Our method obtains at least 8.2\% closer precision to manual evaluation than the held-out evaluation. 

\subsection{Effect of Vetting Strategy}

\begin{figure}[ht]
\centering
\includegraphics[width=0.98\linewidth]{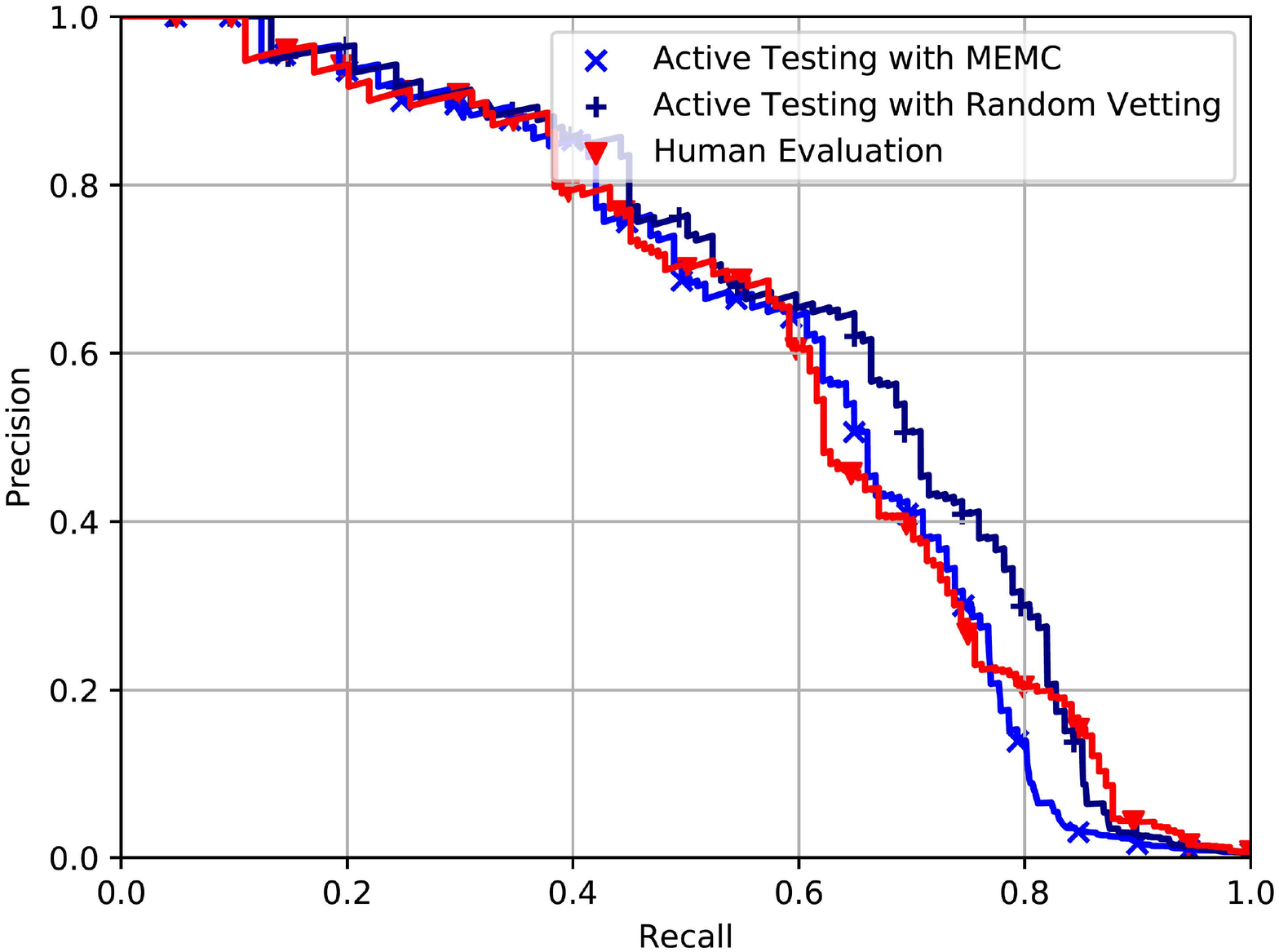}
\caption{The PR curves of PCNN+ATT evaluated with various vetting strategies on NYT-19}
\label{fig:stratergies}
\end{figure}

We compare our MEMC strategy with a random vetting strategy as shown in Figure~\ref{fig:stratergies}. The distance from curves of different vetting strategies to that of human evaluation is 0.176 and 0.284. From the figure, we can conclude that the proposed vetting strategy is much more effective than the random vetting strategy. With the same vetting budget, MEMC gets more accurate  performance estimation at most parts of the range.

\subsection{Re-evaluation of Relation Extractors}
With the proposed performance estimator, we re-evaluate eight up-to-date distantly supervised relation extractors.

\label{sec:supplemental}
\begin{table}[htbp]
\resizebox{\linewidth}{!}{
\centering
\begin{tabular}{cccccc}
\hline
Model & P@100(\%)  & P@200(\%) & P@300(\%) \\
\hline
\citealt{zeng2015distant} & 88.0 & 85.1 & 82.3\\
\citealt{lin2016neural} & 91.2 & 88.9 & 83.8\\
\citealt{liu2017soft} & 94.0 & 89.0 & 87.0\\
\citealt{qin2018robust} & 88.8 & 86.2 & 84.8\\
\citealt{qin2018dsgan} & 87.0 & 83.8 & 80.8\\
\citealt{liu2018neural} & \textbf{95.7} & \textbf{93.4} & \textbf{89.9}\\
BGRU & 94.4 & 89.5 & 84.7\\
BGRU+ATT & 95.1 & 90.1 & 87.1\\
\hline
\end{tabular}}
\caption{The P@N precision of distantly supervised relation extractors on NYT-10. All the methods are implemented with the same framework and running in the same run-time environment.}
\label{table:app_topk}
\end{table}

From Table~\ref{table:app_topk}, we can observe that: 1) The relative ranking of the models according to precision at top $K$ almost remains the same except \citealt{qin2018robust} and \citealt{qin2018dsgan}. Although GAN and reinforcement learning are helpful to select valuable training instances, they are tendentiously to be over-fitted. 2) Most models make the improvements as they mentioned within papers at high confident score interval. 3) BGRU performs better than any other models, while BGRU based method \citealt{liu2018neural} achieves highest precision. More results and discussions can be found in the Appendix.

\section{Conclusion}
\label{sec:conc}

In this paper, we propose a novel active testing approach for distantly supervised relation extraction, which evaluates performance of relation extractors with both noisy data and a few vetted data. Our experiments show that the proposed evaluation method is appropriately unbiased and significant for optimization of distantly relation extraction in future. 

\section{Acknowledgements}
This work is partially supported by Chinese National Research Fund (NSFC) Key Project No. 61532013 and No. 61872239; BNU-UIC Institute of Artificial Intelligence and Future Networks funded by Beijing Normal University (Zhuhai) and AI and Data Science Hub, BNU-HKBU United International College (UIC), Zhuhai, Guangdong, China.

\bibliography{emnlp2020}
\bibliographystyle{acl_natbib}

\appendix

\section{Appendices}
\label{sec:appendix}
\subsection{Logistic Regression}
Here we provide the derivation of Equation.\ref{equ:pzi} in the main paper.

\begin{equation*}
\begin{small}
\begin{aligned}
  p(z'_i|y'_i, s'_i) & = \frac{p(z'_i,  y'_i, s'_i)}{\sum_v p(z'_i=v,y'_i, s'_i)} \\
  & =   \frac{p(z_{jk},  y_{jk}, s_{jk})}{\sum_v p(z_{jk}=v, y_{jk}, s_{jk})} \\
  & = \frac{p(y_{jk}|z_{jk}, s_{jk})p(z_{jk}|s_{jk})}{\sum_v p(y_{jk}|z_{jk}=v,s_{jk})p(z_{jk}=v|s_{jk})} \\
\end{aligned}
\end{small}
\end{equation*}

We assume that given $z_{jk}$, the observed label $y_{jk}$ is conditionally independent of $s_{jk}$, which means $p(y_{jk}|z_{jk}, s_{jk})=p(y_{jk}|z_{jk})$. The expression is simplified to:

\begin{equation*}
\begin{small}
\begin{aligned}
p(z'_i|y'_i, s'_k) = \frac{p(y_{jk}|z_{jk})p(z_{jk}|s_{jk})}{\sum_v p(y_{jk}|z_{jk}=v)p(z_{jk}=v|s_{jk})}
\end{aligned}
\end{small}
\end{equation*}

\subsection{Vetting Strategy}
Here we provide the derivation of Equation.\ref{equ:epz} in the main paper.

\begin{equation*}
\begin{small}
\begin{aligned}
E_{p(z'_i|V)}[\Delta_i(z'_i)] & = p_i\frac{1}{K}|1-p_i|+(1-p_i)\frac{1}{K}|0-p_i| \\
& = \frac{2}{K}p_i(1-p_i)
\end{aligned}
\end{small}
\end{equation*}

\subsection{Experimental result of BGRU+ATT}
\begin{table}[ht]
\centering
\resizebox{\linewidth}{!}{
\begin{tabular}{ccccc}
\hline
Model & Evaluations & P@100 & P@200 & P@300 \\
\hline
\multirow{3}{*}{BGRU+ATT}& Held-out Evaluation  & 82 & 78.5 & 74.3\\
& Our method & 95.2 & 90.1 & 87.1 \\
& Human Evaluation  & 98 & 96& 95 \\
\hline
\end{tabular}}
\caption{The Precision at top K predictions (\%) of BGRU+ATT upon held-out evaluation, our method and human evaluation on NYT-10.}
\label{table:result_topk_bgru_att}
\end{table}

\begin{figure}[ht]
\centering
\centering
\includegraphics[width=0.98\linewidth]{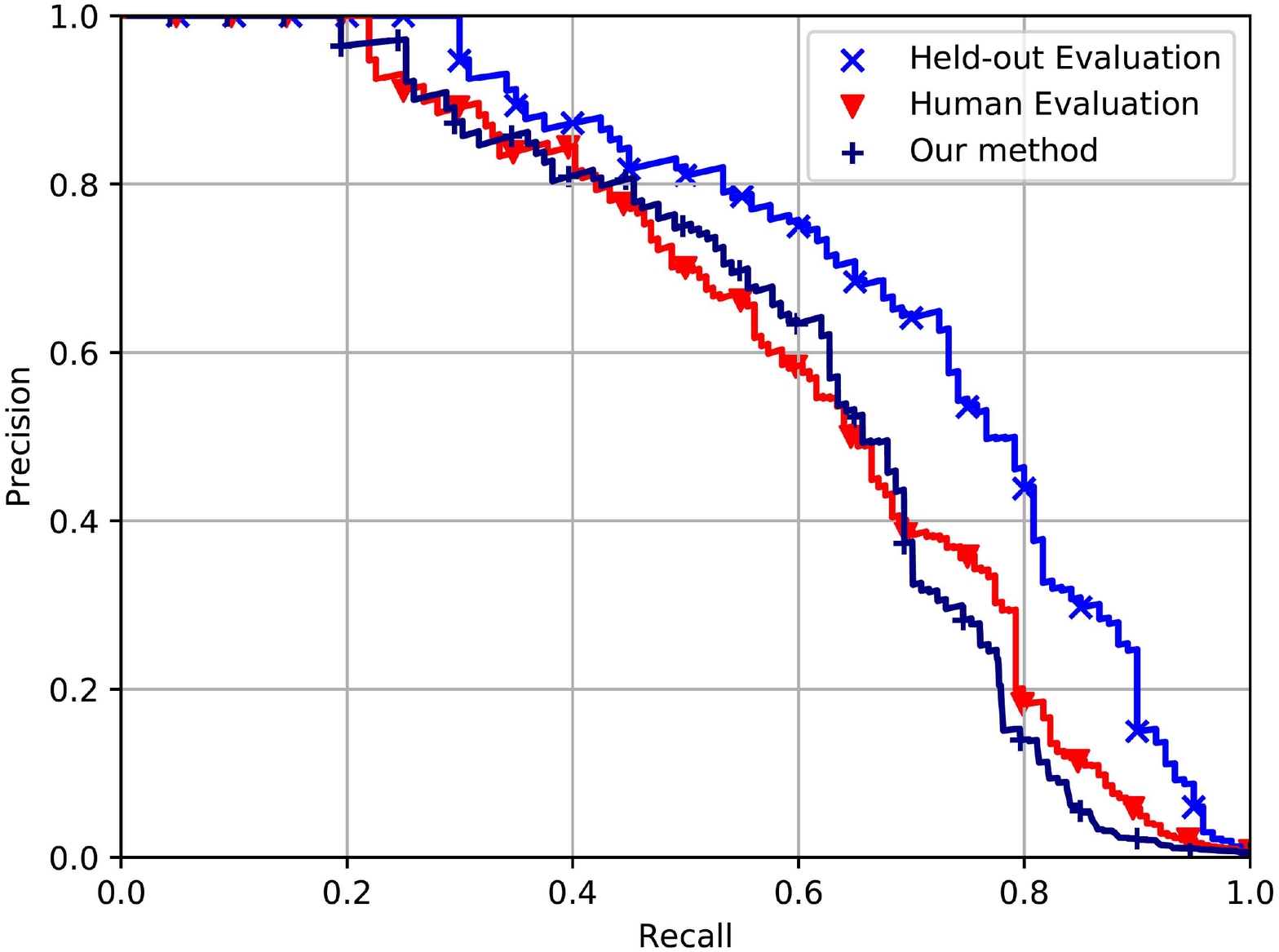}
\caption{The PR curve of BGRU+ATT on NYT-19.}
\label{fig:bgru_PR}
\end{figure}

We also evaluate the performance of BGRU+ATT with held-out evaluation, human evaluation and our method. The results are shown in Table~\ref{table:result_topk_bgru_att}, and Figure~\ref{fig:bgru_PR}. Our method gets the distances 0.15 to the curve of human evaluation while corresponding distances for held-out evaluation is 0.55.

\subsection{The result of different iterations}

\begin{figure}[h]
\centering
\includegraphics[width=0.98\linewidth]{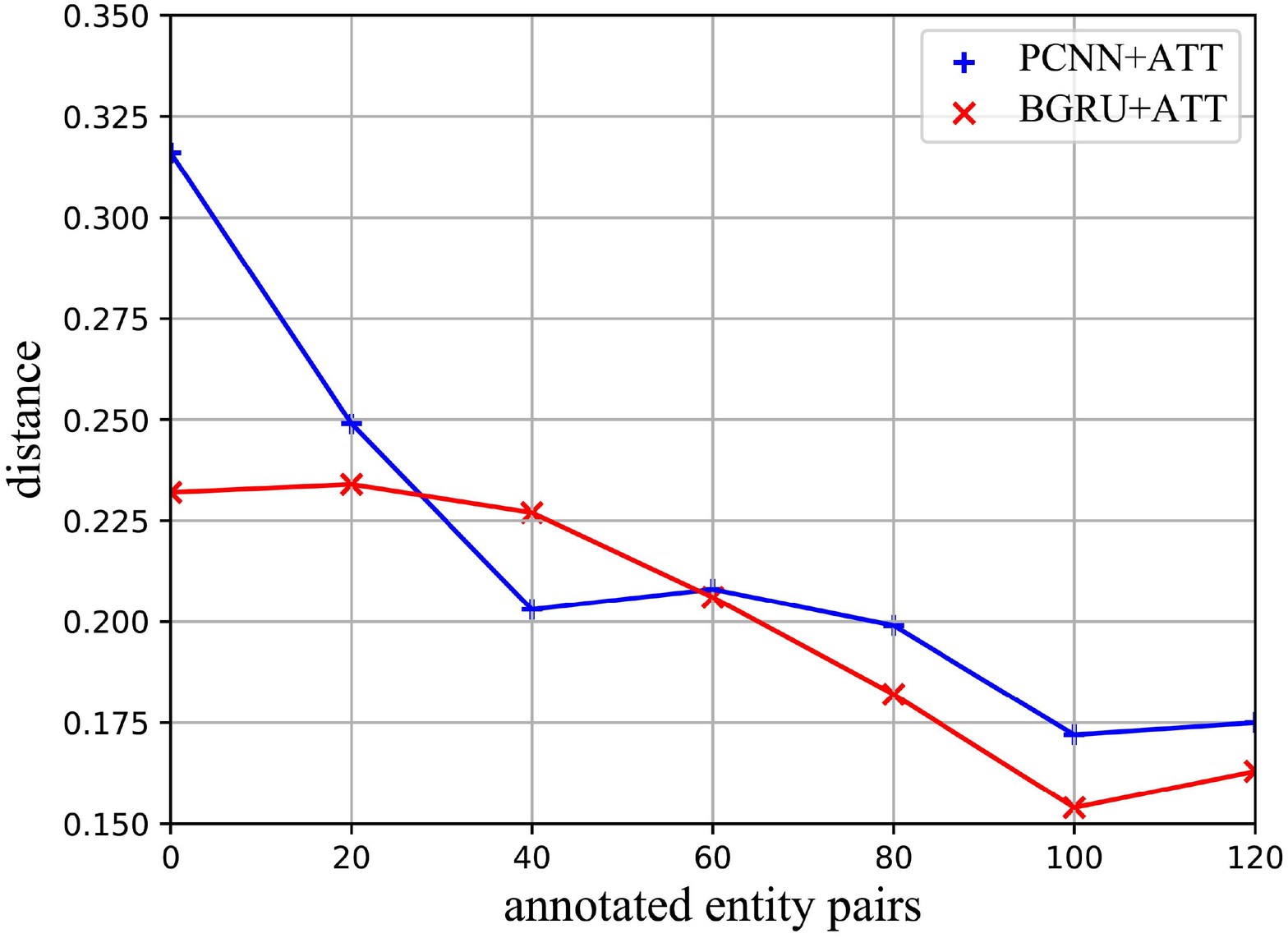}
\caption{The result of different iterations for the active testing algorithm with PCNN+ATT and BGRU+ATT}
\label{fig:iteration}
\end{figure}

We have recorded the distance of different iterations between the curves obtained by our method and manual evaluation in Figure~\ref{fig:iteration}. With the results, we can observe that the evaluation results obtained by our method become closer to human evaluation when the number of annotated entity pairs is less than 100. When the number is more than 100, the distance no longer drops rapidly but begins to fluctuate.

\section{Case Study}

\begin{figure*}[htbp]
\begin{adjustbox}{center}
\includegraphics[width=0.98\linewidth]{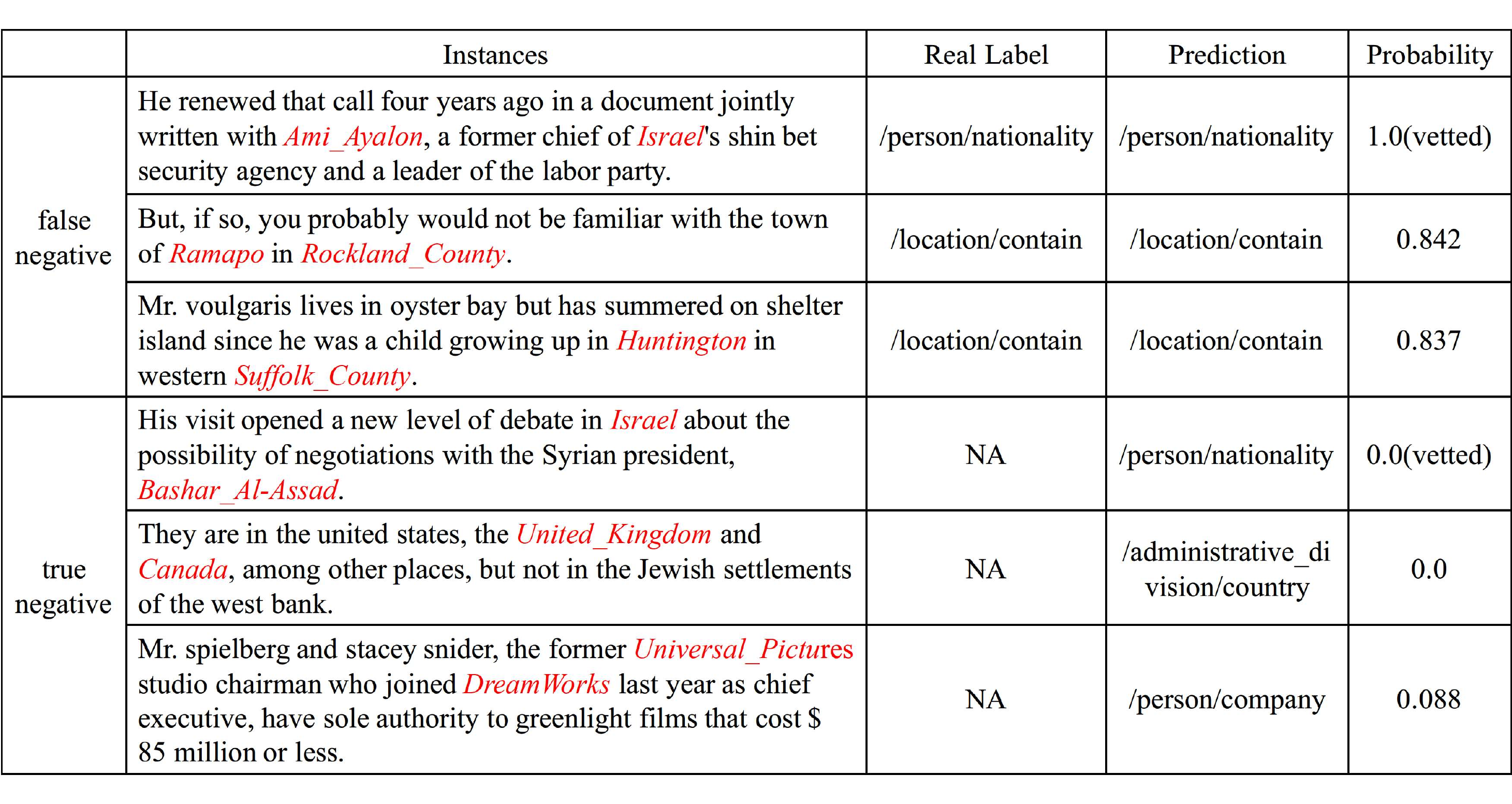}
\end{adjustbox}
\caption{A case study of active testing approach for distantly supervised relation extraction. The entities are labeled in red. $1.0(vetted)$ and $0.0(vetted)$ mean that the entity pair is vetted in our method. }
\label{fig:case_study}
\end{figure*}

We present realistic cases in NYT-10 to show the effectiveness of our method. In Figure~\ref{fig:case_study}, all cases are selected from Top 300 predictions of PCNN+ATT. These instances are all negative instances and has the automatic label $NA$ in NYT-10. In held-out evaluation, relation predictions for these instances are judged as wrong. However, part of them are false negative instances in fact and have the corresponding relations, which cause considerable biases between manual and held-out evaluation. In our approach, those relation predictions for false negative instances are given a high probability to be corrected. At the same time, true negative instances are accurately identified and given a low (near zero) probability.

\section{Re-evaluation Discussion}

The detailed descriptions and discussions of re-evaluation experiments are conducted in this section.

\subsection{Models}
\noindent{\bf PCNN~\cite{zeng2015distant}} is the first neural method used in distant supervision without human-designed features.

\noindent{\bf PCNN+ATT~\cite{lin2016neural}} further integrates a selective attention mechanism to alleviate the influence of wrongly labeled instances. The selective attention mechanism generates attention weights over multiple instances, which is expected to reduce the weights of those noisy instances dynamically. 

\noindent{\bf PCNN+ATT+SL~\cite{liu2017soft}} is the development of PCNN+ATT. To correct the wrong labels at entity-pair level during training, the labels of entity pairs are dynamically changed according to the confident score of the predictive labels. Clearly, this method highly depends on the quality of label generator, which has great potential to be over-fitting.

\noindent{\bf PCNN+ATT+RL~\cite{qin2018robust}} adopts reinforcement learning to overcome wrong labeling problem for distant supervision. A deep reinforcement learning agent is designed to choose correctly labeled instances based on the performance change of the relation classifier. After that, PCNN+ATT is adopted on the filtered data to do relation classification.

\noindent{\bf PCNN+ATT+DSGAN~\cite{qin2018dsgan}} is an adversarial training framework to learn a sentence level true-positive generator. The positive samples generated by the generator are labeled as negative to train the generator. The optimal generator is obtained when the discriminator cannot differentiate them. Then the generator is adopted to filter distant supervision training dataset. PCNN+ATT is applied to do relation extraction on the new dataset.

\noindent{\bf BGRU} is one of recurrent neural network, which can effectively extract global sequence information. It is a powerful fundamental model for wide use of natural language processing tasks.

\noindent{\bf BGRU+ATT} is a combination of BGRU and the selective attention.

\noindent{\bf STPRE~\cite{liu2018neural}} extracts relation features with BGRU. To reduce inner-sentence noise, authors utilize a Sub-Tree Parse(STP) method to remove irrelevant words. Furthermore, model parameters are initialized with a prior knowledge learned from the entity type prediction task by transfer learning.

\begin{figure}[htbp]
\centering
\includegraphics[width=0.98\linewidth]{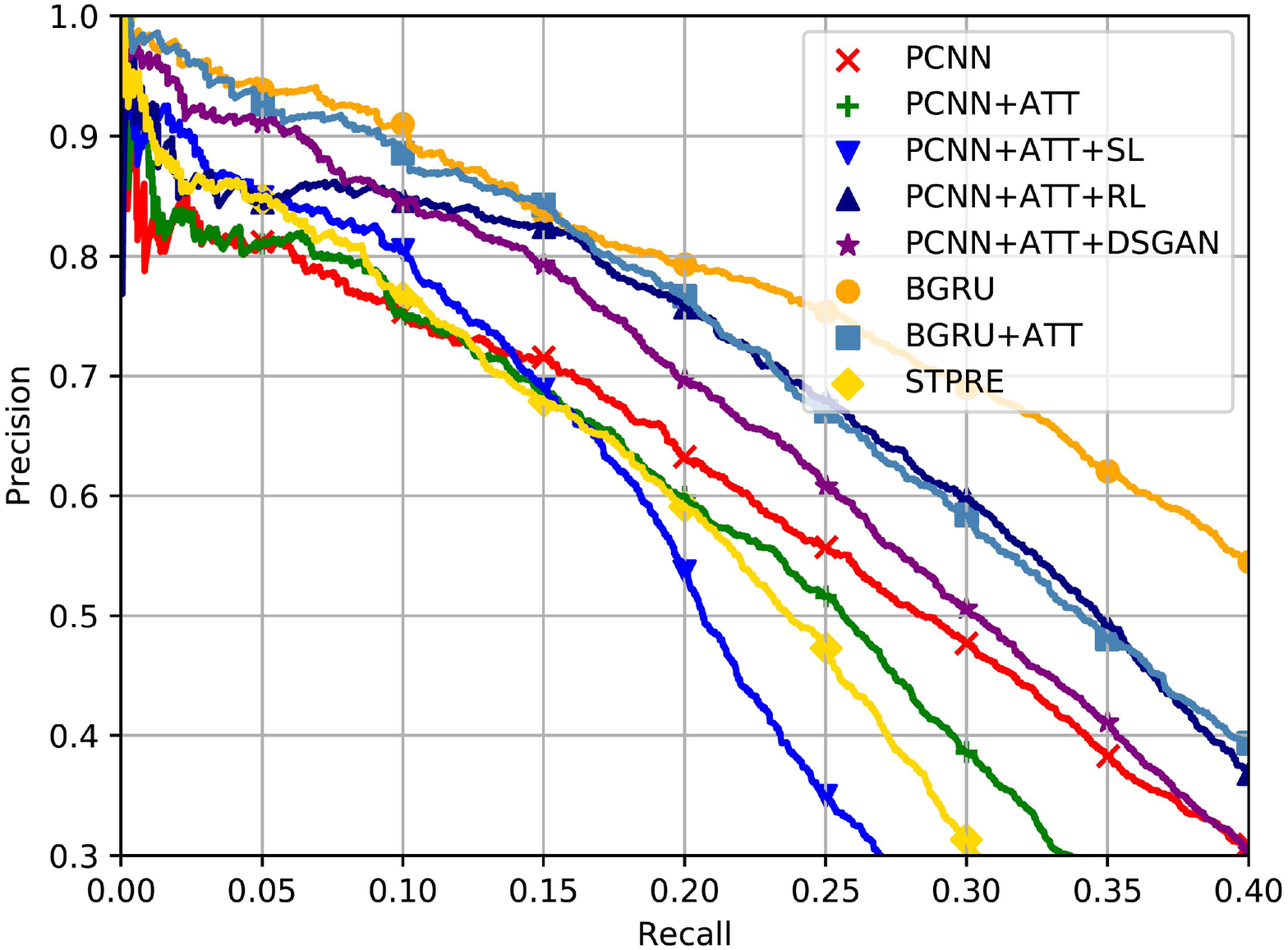}
\caption{PR curve of distantly supervised relation extractors on NYT-10 with the proposed active testing.}
\label{fig:app_PR}
\end{figure}

\subsection{Discussion}

In this section, we additionally provide PR curves to show the performance of baselines. From both Table~\ref{table:app_topk} and Figure~\ref{fig:app_PR}, we are aware of that: 1) The relative ranking is quite different from that on held-out evaluation according to PR curve. 2) The selective attention has limited help in improving the overall performance, even though it may have positive effects at high confident score. 4) The soft-label method greatly improves the accuracy at high confident score but significantly reduces the overall performance. We deduce that it is severely affected by the unbalanced instance numbers of different relations, which will make label generator over-fitting to frequent labels. 4) For the overall performance indicated by PR curves, BGRU is the most solid relation extractor.

\end{document}